\documentclass[]{article}

\usepackage{epsfig}
\usepackage{subcaption}
\usepackage{calc}
\usepackage{amssymb}
\usepackage{amstext}
\usepackage{amsmath}
\usepackage{amsthm}
\usepackage{multicol}
\usepackage{pslatex}

\usepackage{multirow}

\usepackage[%
  natbibapa
]{apacite} 

\usepackage{parskip}
\usepackage{hyperref}
\RequirePackage{doi}

\usepackage{SCITEPRESS}     

\begin{document}

\title{Causal Campbell-Goodhart's law and Reinforcement Learning}

\author{\authorname{Hal Ashton \sup{1}\orcidAuthor{0000-0002-1780-9127}} %
\affiliation{\sup{1}Computer Science, University College London, UK} \email{ucabha5@ucl.ac.uk}}


\keywords{Reinforcement Learning, Goodhart's Law, Campbell's Law, Causal Inference, Cognitive Error}

\abstract{Campbell-Goodhart's law relates to the causal inference error whereby decision-making agents aim to influence variables which are correlated to their goal objective but do not reliably cause it. This is a well known error in Economics and Political Science but not widely labelled in Artificial Intelligence research. Through a simple example, we show how off-the-shelf deep Reinforcement Learning (RL) algorithms are not necessarily immune to this cognitive error. The off-policy learning method is tricked, whilst the on-policy method is not. The practical implication is that naive application of RL to complex real life problems can result in the same types of policy errors that humans make. Great care should be taken around understanding the causal model that underpins a solution derived from Reinforcement Learning. }

\onecolumn \maketitle \normalsize \setcounter{footnote}{0} \vfill

\section{\uppercase{Introduction}}
\label{sec:introduction}

\noindent In many learning tasks, the learning agent has an impact on a stochastic state variable through its previous actions. If when undisturbed, this variable predicts but does not reliably cause something of interest to the agent, the agent might choose actions to target this variable thereby causing suboptimal performance. This is related to Campbell-Goodhart's law in social science which is described as: "When a measure becomes a target, it ceases to be a good measure" \citep{Strathern1997ImprovingSystem}. \citet{Manheim2018CategorizingLaw} describe it occurring "When optimization causes a collapse of the statistical relationship between a goal which the optimizer intends and the proxy used for that goal".  

\noindent In such a situation it is important to have a causal model for the learning task so that the agent is able to correctly separate the effect of its actions on the world, from those which are caused by some other mechanism. According to \cite{Pearl2018TheEffect}, reasoning based on correlations alone is not sufficient to solve certain causal problems. So-called level 2 and 3 problems on their inference ladder require concepts of causality, intervention and counterfactual reasoning. 

\noindent Recently 'Deep' Reinforcement Learning (RL) has had great success in the automated mastery of learning optimal policies to hard problems such as Chess and Go \citep{Silver2017MasteringKnowledge} and a range of computer games starting with Atari \citep{Mnih2015Human-levelLearning} through to more advanced games like StarCraft \citep{Vinyals2019GrandmasterLearning} through the use of Neural networks. All of these problems have environments where there is an effect to the optimizer's actions and yet RL has traditionally avoided discussing causality at all\footnote{For example the canonical text in Reinforcement Learning: \cite{Sutton2018ReinforcementIntroduction} makes no explicit reference to causality throughout the book.}. Are the successes stated above made possible because the problems have straightforward causal dependencies? Perhaps inside the black boxes derived during training, a causal inference technique is found automatically. Else, are RL methods easily confounded by simple causal problems? It seems useful to know for anyone wishing to use RL for real world applications like finance. The inability to explain AI coupled with claims as to its superhuman abilities is termed 'enchanted determinism' in \cite{Campolo2020EnchantedIntelligence}. Over-confidence and an accountability shield are two ill-effects of this phenomenon.

\noindent In this paper I present a toy-problem where an agent is able to alter (or intervene on) a variable that can be otherwise used to predict its reward. This problem is classified as Causal Goodhart by Mannheim and Garrabant (Specifically metric manipulation). It is deliberately simple and it can be solved either analytically or using a number of learning methods without involving neural networks. The motivation is to see whether problems that have an interesting causal structure can be solved through a naive application of existing, off the shelf RL learning algorithms which have had success solving complex problems. 

\section{\uppercase{The Dog Barometer problem}}

\noindent There is a dog living in a house in Scotland that wants to go for a walk. The dog can observe current weather through a window but really needs to know future weather when it is walking.

\noindent The capricious weather of Scotland is either Rain or Sunshine and depends on recent barometric pressure only. 

\noindent The barometric pressure (henceforth just pressure) is either high or low. Future pressure depends only on past pressure. High pressure causes sunshine more often and low pressure causes rain more often. 

\noindent The dog would like to wear its smart coat when it is raining, but not wear it when it is sunny. Once it has committed to its sartorial choice, the dog leaves its house and experiences the weather during its walk. This marks the end of the decision problem for the dog.

\noindent Within the house there is a barometer which measures current pressure and has two states high and low. The dog can see the barometer. The barometer also has a button which the dog can press. The effect is to set its reading to high.\footnote{The barometer use is inspired from a lecture given by Prof Ricardo Silva at UCL} 

\subsection{Causal structure}\label{sec:cause}
High pressure ($P_t=1$) causes a high barometer reading ($B_t=1$) and a high chance of sunshine period ($W_{t+1}=1$).

\noindent Conversely Low pressure ($P_t=1$) causes a low barometer reading ($B_t=1$) and a higher chance of rain next period ($W_{t+1}=1$).

\noindent Touching the barometer causes a high barometer reading ($B_{t+1}=1$) next period regardless of the pressure

\begin{figure}[!ht]
    \centering
    \resizebox{0.5\textwidth}{!}{
    \includegraphics{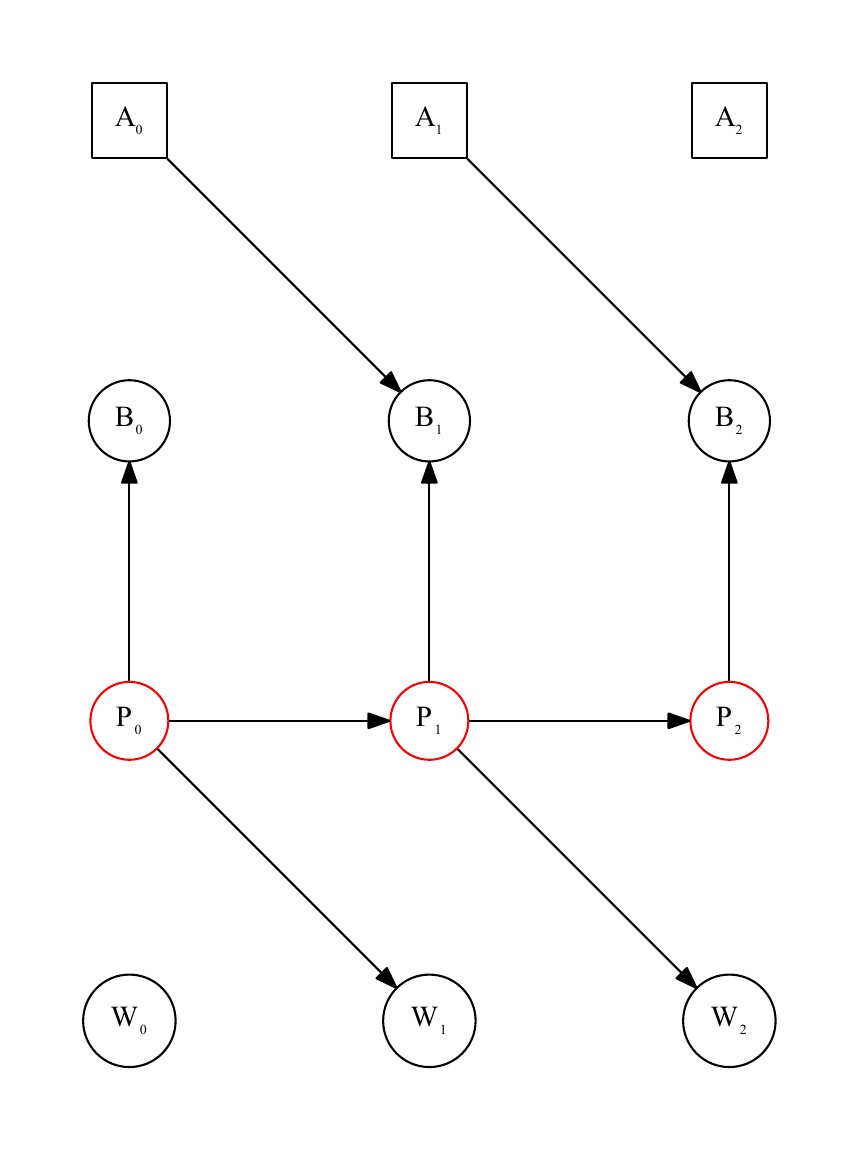}
    }
    \caption{Causal diagram for dog barometer problem shown for 3 periods. $A_t$ variables represent actions, $B_t$ pressure readings from the barometer, $P_t$ the actual pressure, and $W_t$ the current weather. Pressure is hidden to the (canine) observer. The button press makes an intervention on the state of $B_i$, accordingly all incoming arrows into $B_i$ should be deleted, thereby making it useless as an indicator for $P_i$.}
    \label{fig:weather 1}
\end{figure}

\noindent This is summarised by the DAG in Figure \ref{fig:weather 1} limited to three periods. Alternatively it can be written with the following independence statement:

\begin{multline}
        \mathbb{P}(P_t,B_t,W_t|P_{t-1},B_{t-1},W_{t-1},A_{t-1})= \\\mathbb{P}(P_t|P_{t-1}).\mathbb{P}(B_t|P_{t},A_{t-1}).\mathbb{P}(W_t|P_{t-1})
\end{multline}

\noindent In the simplest case, Pressure has no autocorrelation: $\mathbb{P}(P_t|P_{t-1})=\mathbb{P}(P_t)$. This also has the effect of making the Weather variable useless to the dog.

\noindent The effect of the dog pressing the button and setting the Barometer variable $B_t$ is akin to that of an atomic intervention \citep{Pearl2000Causality:Inference}. This removes any arcs from parental nodes leading to $B_t$, meaning that inference about the state of $P_{t}$ is impossible.

\subsection{Problem as a MDP}

We will model this problem as a MDP (Markov Decision Process)\footnote{Assuming that pressure is not observable, it is better modelled as a Partially Observable Markov Decision Process (POMDP), but we will proceed naively '\`a la mode' and ignore the hidden variable.}. 
Time is discretised and indexed by $t$. An MDP is  a tuple $(\mathcal{S},\mathcal{A},R,\mathcal{T},\mathcal{S}_0,\gamma)$ where:
\begin{enumerate}
    \item $\mathcal{S}$ is the set of states. The binary random weather variable be W, with rain denoted $W=0$ and sunshine $W=1$. Pressure variable is P, with a high state denoted $P=1$ and a low state $P=0$. Similarly for Barometer reading  $B$.
    \item $\mathcal{A}$ is the set of actions.  The dog is able to do one of four things at any time period.
        \begin{enumerate}
            \item Do nothing and wait $a_t=w$
            \item Touch the barometer $a_t=m$
            \item Put on a coat and leave their kennel $a_t=c$
            \item Leave the kennel without a coat. $a_t=n$
        \end{enumerate}
    \item $R:\mathcal{S}\times \mathcal{A}:\rightarrow \mathbb{R}$ is the reward function. The dog prefers to go outside in the sun ($r_{nS}$) with no coat, quite likes going outside with a coat in the rain ($r_{cR}$) but dislikes wearing a coat in the sun ($r_{cS}$) and going out in the rain without a coat ($r_{nR}$). These four rewards are ordered $r_{nS}\geq r_{cR} \geq r_{cS} \geq r_{nR} $. An example assignment is shown in table \ref{tab:reward}. 

    \begin{table}[!ht]
    \centering{
    \begin{tabular}{|l|l|l|}
    \hline
            & Rain & Sun \\ \hline
    Coat    & $r_{cR}=4$    & $r_{cS}=-8$  \\ \hline
    No Coat & $r_{nR}=-8$   & $r_{nS}=8$   \\ \hline
    \end{tabular}
    \caption{Rewards for the dog}
    \label{tab:reward}
    }
    \end{table}
    
    \noindent There is also an optional penalty $r_{wait}=-1$ for the dog when it chooses actions $w$ and $m$, which involve it waiting in the house. Whilst the discount factor would seem to have a similar effect, such a penalty is often used in practice to avoid sparse reward signals and help the learning algorithm.
    
    \item $\mathcal{T}(s,a,s')=\mathbb{P}(s'|s,a)$ is the transition function determining the probability of transitioning from state $s$ to $s'$ after taking action $a$. It is shown in tables \ref{tab:transit_p}, \ref{tab:transit_b} and \ref{tab:transit_w}. \begin{table}[!ht]
       \centering
        \begin{tabular}{|l|l|l|}
        \hline
           $\mathbb{P}(P_{t}|P_{t-1})$  & $P_{t-1}=Low$ & $P_{t-1}=High$ \\ \hline
        $P_t=Low$  & $\rho_{LL}$ & $\rho_{LH}=1-\rho_{HH}$  \\ \hline
        $P_t=High$ & $\rho_{HL}=1-\rho_{LL}$ & $\rho_{HH}$  \\ \hline
        
        \end{tabular}
        \caption{$\mathbb{P}(P_t|P_{t-1})$. Pressure transition: When $\rho_{HH}=\rho_{LL}=0.5$ there is no autocorrelation.}
        \label{tab:transit_p}
        
        \end{table}

        \begin{table}[!ht]
        \centering
        \resizebox{0.45\textwidth}{!}{
        \begin{tabular}{|l|l|l|l|l|}
        \hline
        \multirow{0}{*}{$\mathbb{P}(B_t|P_{t},A_{t-1}$)}{} & \multicolumn{2}{l|}{$A_{t-1}=0$} & \multicolumn{2}{l|}{$A_{t-1}=1$} \\ \cline{2-5} 
                          & $P_{t}=Low$        & $P_{t}=High$        & $P_{t}=Low$        & $P_{t}=High$        \\ \hline
        $B_t=Low$             & $\alpha_{L}=0.9$          & $1-\alpha_H$           & 0          & 0           \\ \hline
        $B_t=High$            & $1-\alpha_{L}$          & $\alpha_{H}=0.9$           & 1          & 1           \\ \hline
        \end{tabular}
        }
        \caption{$\mathbb{P}(B_t|P_{t-1},A_{t-1})$ If the button on the barometer has been touched in the previous period ($A_{t-1}=1$) the probability of a high reading is 1. The $\alpha_.$ coefficients  correspond to the accuracy of the barometer.}
        \label{tab:transit_b}
        \end{table}

        \begin{table}[!ht]
        \centering
        \begin{tabular}{|l|l|l|}
        \hline
        $\mathbb{P}(W_t|P_{t-1})$     & $P_{t-1}=Low$ & $P_{t-1}=High$ \\ \hline
        $W_t=Rain$  & $\omega_{RL}=0.9$   & $1-\omega_{SH}$    \\ \hline
        $W_t=Sun$ & $1-\omega_{RL}$   & $\omega_{SH}=0.9$  \\\hline  
        \end{tabular}
        \caption{$\mathbb{P}(W_t|P_{t-1})$. The $\omega_{..}$ coefficients correspond to the capriciousness of the weather.}
        \label{tab:transit_w}
        \end{table}
    \item $S_0\in \mathbb{P}(\mathcal{S})$ is a distribution over states that the process begins at. For the initial states we draw $P_{-1}=H$ with probability 0.5 and generate $P_0$, $B_0$ and $W_0$ according to the conditional distributions in tables \ref{tab:transit_p},\ref{tab:transit_b} and \ref{tab:transit_w} respectively.
    \item $\gamma =0.95$ is a discount factor which reflects how much less the dog values future rewards from present ones.
\end{enumerate}

\noindent The dog must choose a policy function $\pi:\mathcal{S}:\rightarrow \mathcal{A}$ to maximise the following discounted sum of rewards:

\begin{equation}
    \arg_\pi \max E  \big[ \sum_t \gamma^t R(s_t,a_t)  \big| \pi \big]
\end{equation}

\section{Method}
\noindent I implemented the Dog Barometer problem in Open AI gym \footnote{https://gym.openai.com/}. I then tested two Deep RL learning algorithms on this problem using the StableBaselines 3 module which provides a number of implementations of state of the art deep RL algorithms \footnote{https://github.com/DLR-RM/stable-baselines3}. The code for the Dog Barometer environment and tests is available online \footnote{Environment available at \url{https://github.com/yetiminer/dogbarometer/}}.

\noindent For each algorithm I tested the case when pressure is (somehow) visible to the dog as a baseline and when it is not. In both cases the algorithm was trained separately 10 times.

\noindent The first learning algorithm tested was DQN, which was shown to be successful learning how to play Atari games in \cite{Mnih2015Human-levelLearning}. It learns through Q-learning \citep{Sutton2018ReinforcementIntroduction} and approximates the State-action function (aka Q-function) through a neural network. The optimal policy is then the action with the highest value for any state A memory of experiences (termed experience replay) is built up to allow batch updates of the neural networks. This algorithm is classified as off-policy learning, since the algorithm estimates the value of an optimal policy without having to follow that policy during exploration. 

\noindent The second learning algorithm I tested was A2C which evolved from \cite{Mnih2016AsynchronousLearning}. This is an actor-critic method which estimates value and actions functions through neural-networks. It is an on-policy learning method, that is to say the algorithm seeks to improve the policy it is currently following.  

\noindent In both cases, I used the default parameters according to StableBaselines 3. In particular all the neural networks were two layered, feed-forward perceptrons of 64 neurons each with $Tanh$ activations.

\noindent The A2C algorithm was trained for 20,000 episodes and the DQN algorithm, being less efficient was trained for 100,000. These figures were chosen for sufficient convergence properties. Default settings from the StableBaselines3 \footnote{\url{https://github.com/DLR-RM/stable-baselines3}} module were used for both algorithms. The resultant strategies were evaluated over 10,000 episodes. 

\section{Results}
\noindent  In Experiment 1, pressure has no auto-correlation: $\rho_{LL}=\rho_{HH}=0.5$. This makes the strategy of waiting for high pressure or a high barometer reading the most efficient. We denote this strategy $\Pi_{nw}$. Table \ref{tab:exp1_results} shows the results of the different training methods. When Pressure is visible, the optimal strategy is recovered by both algos though some of the time A2C converges on $\Pi_{nc}$ - wear or don't wear coat according to barometer. When pressure is hidden, the A2C algorithm successfully finds the optimal strategy $\Pi_{nw}$ on every occasion. The DQN algorithm always converges on $\Pi_{nb}$; the naive strategy that involves pressing the barometer button if the barometer is initially low and going outside without a coat if the barometer is high.

\begin{table}[!ht]
\centering
\resizebox{0.45\textwidth}{!}{%
\begin{tabular}{|l|l|l|l|l|l|}
\hline
\multirow{0}{*}{Exp series} & \multirow{0}{*}{Pressure Hidden} & \multirow{0}{*}{Mean Reward} & \multicolumn{3}{l|}{Strategy count} \\ \cline{4-6} 
 &  &  & $\Pi_{nc}$ & $\Pi_{nw}$ & $\Pi_{nb}$ \\ \hline
A2C &  & 4.80 & 3 & 7 &  \\ \hline
A2C$_H$ & TRUE & 4.15 &  & 10 &  \\ \hline
DQN &  & 5.39 &  & 10 &  \\ \hline
DQN$_H$ & TRUE & 2.05 &  &  & 10 \\ \hline
\end{tabular}%
}
\caption{DQN trained dogs are consistently fooled into pressing the barometer - strategy $\Pi_{nb}$ whilst the A2C algorithm correctly recovers the optimal strategy - $\Pi_{nc}$}
\label{tab:exp1_results}
\end{table}

\begin{table}[!ht]
\centering
\resizebox{0.45\textwidth}{!}{%
\begin{tabular}{|l|l|l|l|l|l|l|}
\hline
\multirow{0}{*}{Exp series} & \multirow{0}{*}{Pressure Hidden} & \multirow{0}{*}{Mean Reward} & \multicolumn{4}{l|}{Strategy count} \\ \cline{4-7} 
 &  &   & $\Pi_{nc}$ & $\Pi_{nb}$ & $\Pi_{nwc}$ & $\Pi_{nbb}$ \\ \hline
A2C E2 &  & 4.58   & 10 &  &  &  \\ \hline
A2C E2 H & TRUE & 3.58   &  &  & 10 &  \\ \hline
DQN E2 &  & 4.60   & 10 &  &  &  \\ \hline
DQN E2 H & TRUE & 0.87   &  & 8 &  & 2 \\ \hline
\end{tabular}%
}
\caption{Strategies found when weather is auto-correlated }
\label{tab:exp2_results}
\end{table}

\noindent In Experiment 2 $\rho_{LL}=\rho_{HH}=0.75$. That is to say pressure has auto-correlation - the probability of maintaining the same level between periods is 0.75. This now makes waiting for high pressure less desirable. It also makes the weather variable useful as an indicator independent to the barometer for the previous state of pressure. Table \ref{tab:exp2_results} shows the results of the different training methods. The A2C algorithm successfully finds the optimal strategy $\Pi_{nwc}$ on every occasion. This is the strategy where the Dog exits the house with or without a coat depending on the barometer unless the barometer reads low and the weather is fine, in which case the dog will wait a period. This balances the chance of a misreading from the barometer, the penalty of going out with a coat when the weather is sunny.
The DQN algorithm mostly converges on $\Pi_{nb}$; the sub-optimal strategy that involves pressing the barometer button if the barometer is initially low but also $\Pi_{nbb}$ which involves pressing the barometer on every occasion except when the barometer and the weather agree on a high/sun reading. This is an improvement on $\Pi_{nb}$ since the weather is being used as an indicator though is still not efficient.

\section{Discussion}
\noindent In our experiments we saw that the DQN method of training consistently led to the naive strategy of pressing the barometer to 'cause' high pressure which would cause desirable sunny weather. In contrast A2C avoids this pitfall and consistently finds an optimal strategy.

\noindent I hypothesise that this could be due to two related features of DQN; Experience replay and Off-policy updating. Experience replay consists of the chunking experience which is subsequently sampled in batches to update the neural network that estimates the state-action value (Q-value) of each state. Because prior-actions are not saved in this memory, the distribution of rewards is not separated between those where the button has been pressed and those where it has not. High reward signals from not wearing a coat after reading a legitimately high barometer reading are mixed with the disappointing ones of pressing the barometer and exiting without a coat. \citet{Bareinboim2016CausalProblem} call this the data-fusion problem. Secondly DQN is an off-policy learning method - all policies are updated during learning not just the policy that the learner is currently following. Again this would seem to mean that the feedback from optimal policies where the button is not pressed is also credited to policies where the button is pressed. 

\noindent In contrast A2C does successfully navigate the Dog-Barometer. This is a surprise given the inadequacy of the state signal in its ability to show when the barometer is behaving properly. On reflection I think this might be because this method of learning is \emph{'on-policy'}. Learning only occurs on a policy which is currently being used by the learner. Since the dynamics of this environment are dependent on the action-history, and in this example, the policy encodes action history, the learner is not tripped up as easily. 

\noindent In all cases I used the default 2 layer 64 neuron feed forward MLP. It would be useful to try a recurrent neural network like an LSTM \cite{GoodfellowBook2016} to see how the results changed.

\section{Related work}

\noindent Causality within Reinforcement Learning is beginning to receive mainstream attention. An introduction to the subject is given by \cite{Bareinboim2020TowardsCRL} and the associated website\footnote{\url{https://crl.causalai.net/} Accessed August 2020.}. \cite{Guo2018AMethods} provide a more general survey of learning causality from data. The general issues surrounding the replay-memory of DQN erroneously mixing data generated under different policies are explored in \cite{Bareinboim2016CausalProblem}.

\noindent Motivated from biological/psychological perspectives \cite{Gershman2015ReinforcementModels} places causal knowledge in model-based and model free RL and provides a novel, simple taxonomy to identify where causal considerations can come into RL. Gershman highlights research that suggest model and model-free reasoning exists within the brain and discusses how they interact with reference to the Dyna architecture of Sutton and Barto\cite{Sutton1990Integrated7th}. 

\noindent \cite {Buesing2019WouldaSearch} present a counterfactual learning technique in the context of POMDPS (Partially observable MDPs). They show how a model based RL approach to POMDPs can be cast in the language of a SCM (Structural Causal Model - see \cite{Pearl2000Causality:Inference}). This is important since SCMs are the principal tool through which causal research has progressed. Counterfactual reasoning is strongly related with off-policy learning methods since it considers the results of actions not taken under the policy used to generate the experience. In contrast to their use of an SCM to generate counter-factual data to search for better policies, for a model-free setting, they observe that sampling from experience can in certain circumstances be high variance or even useless when evaluating different policies. This echoes the poor performance of DQN in our experiment.

\noindent The subject of unobserved confounding in MDPs (termed MDPUCs) is studied in \cite{Zhang2016MarkovApproach}. The authors point out that MDPUCs are quite separate from POMDPS. They observe that, to date, MDP learning algorithms do not differentiate between passive data collection and data collection after actions (interventions). The authors go on to show that standard MDP techniques as used in RL are not guaranteed to converge to optimal policies and present a method using counterfactual analysis to improve upon existing learning algorithms. 

\noindent Model based Reinforcement Learning (MBRL) is an area of RL research typically separate from the RL mainstream which is model-free. In it an agent builds a representation of the world from observational data in order to predict future observations and thereby plan future actions and optimise a policy. \cite{Rezende2020CausallyLearning} study the problem of causal errors arising from partially modelled environments in RL and show why they occur and propose a way of mitigating them. They illustrate the problem with a simple MDP called 'FuzzyBear' and explicitly relate it to causal reasoning's concepts of interventions, backdoors and frontdoors taken from \cite{Pearl2000Causality:Inference}.  

\noindent Goodhart's Law in economics \citep{Goodhart1984ProblemsExperience} is analogous to Campbell's law \citep{Campbell1979AssessingChange} in social science. Since both were originally communicated at around the same point in time, I thought it was important to combine the two to aid communication, as espoused in \cite{Rodamar2018ThereGoodhart}. This article is concerned with one variant of Goodhart-Campbell termed 'causal' in \cite{Manheim2018CategorizingLaw}. Whilst there is a growing knowledge base on the failure cases of AI (see for example \cite{Lehman2020TheCommunities}) the relevant ones are most often examples of what Manheim and Garrabant term 'Shared-Cause' Campbell-Goodhart, which is an example of objective misalignment - the AI learns to maximise the metric but not the ultimate goal of the programmer. To my knowledge, AI research has not explicitly identified Causal-Goodhart effects when discussing failure cases. One potential example of Causal Campbell-Goodhart is to be found in \cite{Ha2018WorldModels}, where the AI which learns an internal model of Doom - a computer game. On occasion it would learn strategies that would work in its model, but would fail when used on the actual computer game. 

\section{Conclusion}
\noindent Reinforcement learning has had great success learning optimal policies in a variety of game settings, often exceeding human competence in Go, Chess, Atari games etc. It is tempting to apply this method to more complex problems where there are hidden variables, stochastic outcomes and non-trivial causal structures and hope for the best. This may lead to disappointment and more seriously, bad policies being enacted. To date, RL Research has given very little consideration to causal issues, often because the causal mechanisms in the canonical test cases are straightforward. By applying RL as is to more complex problems, there is an implicit assumption that the neural networks used in deep RL, are able to figure out the problem of understanding causality just as well as they can learn to decode the raw vision data fed to them in Atari Games as in Mnih et al. and turn it into winning strategies. \cite{Pearl2000Causality:Inference} argues that certain problems cannot be solved using correlation based statistics alone; to progress to the second rung of his causation ladder, interventions need to be made. RL is a learning framework which naturally performs interventions on the environment it seeks to learn about yet its tools do automatically account for the mathematical implications of making interventions. 

\noindent The dog barometer problem that I present here is deliberately simple and the state-space given to the learner is not ideal for a learning algorithm. RL Methods do exist for settings with hidden variables and I have not used them here. Expanding the state space to include previous state values and actions may solve the problem. However I do think that the problems in RL raised by dog barometer are not simply the result of a straw-man argument. The presence of hidden variables in real life learning applications is almost certain as is the existence of non-trivial causal structures whose effect may linger over arbitrarily long timescales thereby negating the efficacy of adding more history. Every model of a real problem will be misspecified to some extent; it is important to understand when and why this matters. The fact that the cognitive error is sufficiently common in social science to be named Goodhart's Law is a good indicator that this is a policy failure case which is likely to appear again and again in real life applications of RL. In defence of RL, the performance of the A2C algorithm even in the face of such misspecification is very promising and warrants further investigation to see whether this is consistent or an artefact of the environment. 

\noindent Finally, I would like to continue to build an open library of causal problems which new RL algorithms can be benchmarked against. Such an approach using the OpenAI interface has already benefited RL and I think such a library will help widen the audience of Causal RL to general RL researchers. In parallel it would be useful to begin to build a taxonomy of cognitive errors that AI suffers from, starting by investigating whether others, similar to Campbell-Goodhart can be recreated with RL.

\section*{\uppercase{Acknowledgements}}
\noindent This work is supported by an EPSRC PhD studentship.


 \bibliographystyle{apacite}
{\small
\bibliography{references}}


\end{document}